\documentclass[11pt,a4paper]{article}
\usepackage[hyperref]{emnlp2018}
\usepackage{CJK}
\usepackage{times}
\usepackage{latexsym}
\usepackage{amssymb,mathtools}
\usepackage{url}
\usepackage{comment}
\usepackage{graphicx}
\usepackage{paralist}
\usepackage{subcaption}
\usepackage{multirow}
\usepackage{algorithm}
\usepackage{algpseudocode}
\usepackage{tablefootnote}

\algnewcommand{\Or}{\textbf{or}}
\algnewcommand{\And}{\textbf{and}}

\aclfinalcopy 
\def\aclpaperid{517} 

\title{CLUSE: Cross-Lingual Unsupervised Sense Embeddings}

\author{Ta-Chung Chi and Yun-Nung Chen\\
National Taiwan University, Taipei, Taiwan \\ {\tt r06922028@ntu.edu.tw\quad y.v.chen@ieee.org}}

\begin{document}
\maketitle
\begin{abstract}
This paper proposes a modularized sense induction and representation learning model that jointly learns bilingual sense embeddings that align well in the vector space, where the cross-lingual signal in the English-Chinese parallel corpus is exploited to capture the collocation and distributed characteristics in the language pair.
The model is evaluated on the Stanford Contextual Word Similarity (SCWS) dataset to ensure the quality of monolingual sense embeddings.
In addition, we introduce Bilingual Contextual Word Similarity (BCWS), a large and high-quality dataset for evaluating cross-lingual sense embeddings, which is the first attempt of measuring whether the learned embeddings are indeed aligned well in the vector space.
The proposed approach shows the superior quality of sense embeddings evaluated in both monolingual and bilingual spaces.\footnote{The code and dataset are available at \url{http://github.com/MiuLab/CLUSE}.}
\end{abstract}

\section{Introduction}
Word embeddings have recently become the basic component in most NLP tasks for its ability to capture semantic and distributed relationships learned in an unsupervised manner.
The higher similarity between word vectors can indicate similar meanings of words.
Therefore, embeddings that encode semantics have been shown to serve as the good initialization and benefit several NLP tasks.
However, word embeddings do not allow a word to have different meanings in different contexts, which is a phenomenon known as polysemy.
For example, ``\emph{apple}'' may have different meanings in \emph{fruit} and \emph{technology} contexts.
Several attempts have been proposed to tackle this problem by inferring multi-sense word representations~\cite{reisinger2010multi,neelakantan2014efficient,li2015multi,lee2017muse}.

These approaches relied on the ``one-sense per collocation'' heuristic~\cite{yarowsky1993one}, which assumes that presence of nearby words correlates with the sense of the word of interest.
However, this heuristic provides only a weak signal for discriminating sense identities, and it requires a large amount of training data to achieve competitive performance.

Considering that different senses of a word may be translated into different words in a foreign language, \citet{guo2014learning} and \citet{vsuster2016bilingual} proposed to learn multi-sense embeddings using this additional signal.
For example, ``\emph{bank}'' in English can be translated into \emph{banc} or \emph{banque} in French, depending on whether the sense is financial or geographical.
Such information allows the model to identify which sense a word belongs to.
However, the drawback of these models is that the trained foreign language embeddings are not aligned well with the original embeddings in the vector space.

This paper addresses these limitations by
proposing a bilingual modularized sense induction and representation learning system.
Our learning framework is the first pure sense representation learning approach that allows us to utilize two different languages to disambiguate words in English.
To fully use the linguistic signals provided by bilingual language pairs, it is necessary to ensure that the embeddings of each foreign language are related to each other (i.e., they align well in the vector space). 
We solve this by proposing an algorithm that jointly learns sense representations between languages.
The contributions of this paper are four-fold:
\begin{compactitem}
\item We propose the first system that maintains purely sense-level cross-lingual representation learning with linear-time sense decoding.
\item We are among the first to propose a single objective for modularized bilingual sense embedding learning.
\item We are the first to introduce a high-quality dataset for directly evaluating bilingual sense embeddings.
\item Our experimental results show the state-of-the-art performance for both monolingual and bilingual contextual word similarities.
\end{compactitem}

\section{Related Work}
There are a lot of prior works focusing on representation learning, while this work mainly focuses on bridging the work about sense embeddings and cross-lingual embeddings and introducing a newly collected bilingual data for better evaluation.

\vspace{-1mm}
\paragraph{Sense Embeddings}
\citet{reisinger2010multi} first proposed multi-prototype embeddings to address the lexical ambiguity when using a single embedding to represent multiple meanings of a word.
\citet{HuangEtAl2012,neelakantan2014efficient,li2015multi,bartunov2016breaking} utilized neural networks as well as the Bayesian non-parametric method to learn sense embeddings.
\citet{lee2017muse} first utilized a reinforcement learning approach and proposed a modularized framework that separates learning of senses from that of words.
However, none of them leverages the bilingual signal, which may be helpful for disambiguating senses.

\vspace{-1mm}
\paragraph{Cross-Lingual Word Embeddings}
\citet{klementiev2012inducing} first pointed out the importance of learning cross-lingual word embeddings in the same space and proposed the cross-lingual document classification (CLDC) dataset for extrinsic evaluation.
\citet{gouws2015bilbowa} trained directly on monolingual data and extracted a bilingual signal from a smaller set of parallel data.
\citet{kovcisky2014learning} used a probabilistic model that simultaneously learns alignments and distributed representations for bilingual data by marginalizing over word alignments.
\citet{hermann2014multilingual} learned word embeddings by minimizing the distances between compositional representations between parallel sentence pairs.
\citet{vsuster2016bilingual} reconstructed the bag-of-words representation of semantic equivalent sentence pairs to learn word embeddings.
\citet{shi2015learning} proposed a training algorithm in the form of matrix decomposition, and induced cross-lingual constraints for simultaneously factorizing monolingual matrices.
\citet{luong2015bilingual} extended the skip-gram model to bilingual corpora where contexts of bilingual word pairs were jointly predicted.
\citet{wei2017variational} proposed a variational autoencoding approach that explicitly models the underlying semantics of the parallel sentence pairs and guided the generation of the sentence pairs.
Although the above approaches aimed to learn cross-lingual embeddings jointly, they fused different meanings of a word in one embedding, leading to lexical ambiguity in the vector space model.

\vspace{-1mm}
\paragraph{Cross-Lingual Sense Embeddings}
\citet{guo2014learning} adopted the heuristics where different meanings of a polysemous word usually can be represented by different words in another language and clustered bilingual word embeddings to induce senses.
\citet{vsuster2016bilingual} proposed an encoder, which uses parallel corpora to choose a sense for a given word, and a decoder that predicts context words based on the chosen sense.
\citet{bansal2012unsupervised} proposed an unsupervised method for clustering the translations of a word, such that the translations in each cluster share a common semantic sense.
\citet{upadhyay2017beyond} leveraged cross-lingual signals in more than two languages.
However, they either used pretrained embeddings or learned only for the English side, which is undesirable since cross-lingual embeddings shall be jointly learned such that they aligned well in the embedding space.

\vspace{-2mm}
\paragraph{Evaluation Datasets}
Several datasets can be used to justify the performance of learned sense embeddings. \citet{HuangEtAl2012} presented SCWS, the first and only dataset that contains word pairs and their sentential contexts for measuring the quality of sense embeddings. 
However, it is a monolingual dataset constructed in English, so it cannot evaluate cross-lingual semantic word similarity.
On the other hand, while \citet{camacho2017semeval} proposed a cross-lingual semantic similarity dataset, it ignored the contextual words but kept only word pairs, making it impossible to judge sense-level similarity.
In this paper, we present an English-Chinese contextual word similarity dataset in order to benchmark the experiments about bilingual sense embeddings.

\begin{figure*}
  \centering
  \includegraphics[width=\linewidth]{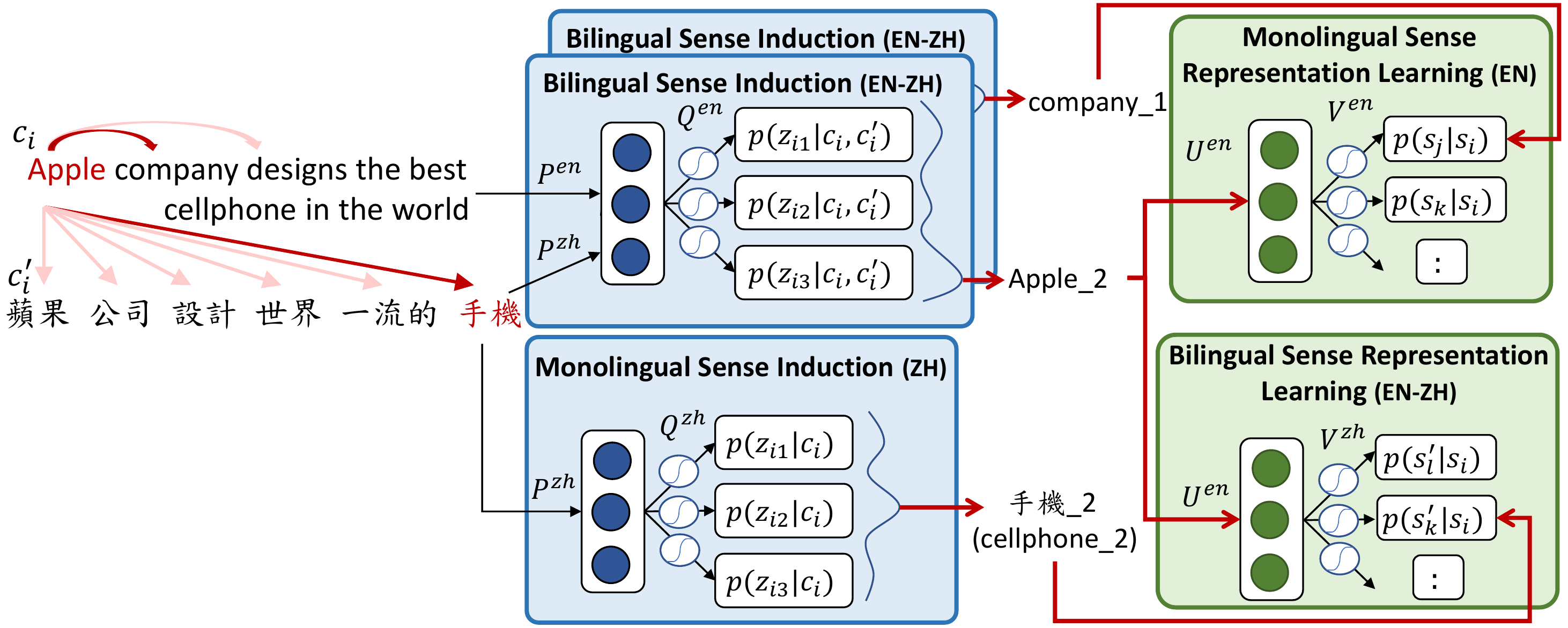}
  \caption{Sense induction modules decide the senses of words, and two sense representation learning modules optimize the sense collocated likelihood for learning sense embeddings within a language and between two languages. Two languages are treated equally and optimized iteratively.}
  \label{fig:model}
\end{figure*}

\section{CLUSE: Cross-Lingual Unsupervised Sense Embeddings}
Our proposed model borrows the idea about modularization from \citet{lee2017muse}, which treats the sense induction and representation modules separately to avoid mixing word-level and sense-level embeddings together.

Our model consists of four different modules illustrated in Figure~\ref{fig:model}, where sense induction modules decide the senses of words, and two sense representation learning modules optimize the sense collocated likelihood for learning sense embeddings within a language and between two languages in a joint manner.
All modules are detailed below.

\subsection{Notations}
We denote our parallel corpus without word alignment $C$, where $C^{en}$ is for the English part and $C^{zh}$ is for the Chinese part.
Our English vocabulary is $W^{en}$ and Chinese vocabulary is $W^{zh}$.
Moreover, $C^{en}_t$ and $C^{zh}_t$ are the $t$-th sentence-level parallel sentences in English and Chinese respectively. 
In the following sections, we treat English as the major language and Chinese as an additional bilingual signal, while their roles can be mutually exchanged.
Specifically, English and Chinese iteratively become the major language during the training procedure.

\subsection{Bilingual Sense Induction Module}
The bilingual sense induction module takes a parallel sentence pair as input and determines which sense identity a target word belongs to given the bilingual contextual information. 
Formally, for the $t$-th English sentence $C^{en}_t$, we aim to decode the most probable sense $z_{ik}\in Z_i$ for the $i$-th word $w_i\in W^{en}$ in $C^{en}_t$, where $Z_i$ is the set of sense candidates for $w_i$ and $1\leq k\leq |Z_i|$. 
We assume that the meaning of $w_i$ can be determined by its surrounding words, or the so-called local context, $c_i=\{w_{i-m},\cdots,w_{i+m}\}$, where $m$ is the size of context window.

Aside from monolingual information, it is desirable to exploit the parallel sentences as additional bilingual contexts to enable cross-lingual embedding learning.
Note that word alignment is not required in this work, so we consider the whole parallel bilingual sentence during training.
Considering training efficiency, 
we sample $M$ words in the parallel bilingual sentence with their original relative order or pad it to $M$ for those shorter than $M$.
Formally, given the $t$-th parallel bilingual sentence $C^{zh}_t$, the bilingual context of $w_i$ is therefore $c'_i=\{w'_0, \cdots ,w'_{M-1}\}$ and $w'\in W^{zh}$.

To ensure efficiency, continuous bag-of-words (CBOW) model is applied, where it takes word-level input tokens and outputs sense-level identities.
Specifically, given an English \emph{word} embedding matrix $P^{en}$, the local context can be modeled as the average of word embeddings
from its context, $\frac{1}{|c_i|}\sum_{w_j\in c_i} P^{en}_j$.
Similarly, we can model the bilingual contextual information given Chinese word embedding matrix $P^{zh}$ using the CBOW formulation and obtain $\frac{1}{M}\sum_{w'_j\in c'_i}P^{zh}_j$.
We linearly combine the contextual information from different languages as:
\begin{equation} \label{eq:1}
\bar{C}=\alpha\cdot \frac{1}{|c_i|}\sum_{w_j\in c_i} P^{en}_j+(1-\alpha)\cdot \frac{1}{M}\sum_{w'_j\in c'_i}P^{zh}_j.
\end{equation}

The likelihood of selecting each sense identity $z_{ik}$ for $w_i$ can be formulated in the form of \emph{Bernoulli} distribution with a sigmoid function $\sigma(\cdot)$:
\begin{equation}\label{eq:2}
  p(z_{ik}\mid c_i, c'_i)= \sigma((Q^{en}_{ik})^T\bar{C}),
\end{equation}
where $Q^{en}$ is a 3-dimensional tensor with each dimension denotes $W^{en}$, $z_{ik}$ for a specific word $i$ in $W^{en}$, and the corresponding latent variable, respectively. Therefore, $Q^{en}_{ik}$ will retrieve the latent variable of $k$-th sense of $i$-th English word.
Finally, we can induce the sense identity, $z^*_{ik}$,
given the contexts of a word $w_i$ from different languages, $c_i$ and $c'_i$.
\begin{equation}\label{eq:3}
  z_{ik}^* = \arg\max_{z_{ik}} {p(z_{ik}\mid c_i, c'_i)}
\end{equation}
In order to allow the module to explore other potential sense identities, we apply an $\epsilon$-greedy algorithm~\cite{mnih2013playing} for exploration in the training procedure.

\subsection{Monolingual Sense Induction Module}
This module is the degraded version of bilingual sense induction module when $\alpha=1$, which occurs where \emph{no} parallel bilingual signal exists.
In other words, every bilingual sense induction module will experience the degradation during the training process presented in Algorithm \ref{algo:learning}.
The only difference is that it cannot access the bilingual information.
The purpose of this module is to maintain the stability of sense induction and to decode the sampled bilingual sense identity which will later be used in the bilingual sense representation learning module.
As shown in Figure~\ref{fig:model}, given the monolingual context of a word, this module selects its sense identity using (\ref{eq:2}) and (\ref{eq:3}) with $\alpha=1$.

\subsection{Monolingual Sense Representation Learning Module}
Given the decoded sense identities from the sense induction module, the skip-gram architecture~\cite{mikolov2013distributed} is applied considering that it only requires two decoded sense identities for stochastic training.
We first create an input English sense representation matrix $U^{en}$ and an English collocation estimation matrix $V^{en}$ as the learning targets.
Given a target word $w_i$ and its collocated word $w_j$ in the $t$-th English sentence $C^{en}_t$, we map them to their sense identities as $z^*_{ik}=s_i$ and $z^*_{jl}=s_j$ by the sense induction module and maximize the sense collocation likelihood.
The skip-gram objective can be formulated as $p(s_j\mid s_i)$:
\begin{equation}
\label{eq:4}
    p(s_j\mid s_i)=
    \frac{\exp((U^{en}_{s_i})^TV^{en}_{s_j})}
    {\sum_{s_k}{\exp((U^{en}_{s_i})^T V^{en}_{s_k})}},
\end{equation}
where $s_k$ iterates over all possible English sense identities in the denominator.
This formulation shares the same architecture as skip-gram but extends to rely on senses.
Note that the Chinese sense representation learning module is built similarly.

\subsection{Bilingual Sense Representation Learning Module}
To ensure sense embeddings of two different languages align well, we hypothesize that the target sense identity $s_i$ not only predicts the sense identity $s_j$ of $w_j$ in $C^{en}_t$ but also one sampled sense identity $s'_l$ of $w'_l$ from the parallel sentence $C^{zh}_t$, where $s'_l$ is decoded by the Chinese monolingual sense induction module.
Specifically, the \emph{bilingual} skip-gram objective can be formulated using the English sense embedding matrix $U^{en}$ and the bilingual collocation estimation matrix $V^{zh}$ as:
\begin{equation}\label{eq:5}
p(s'_l\mid s_i)=\frac{\exp((U^{en}_{s_i})^TV^{zh}_{s'_l})}{\sum_{s'_k} \exp((U^{en}_{s_i})^TV^{zh}_{s'_k})},
\end{equation}
where $s'_k$ iterates over all possible Chinese sense identities in the denominator.

\subsection{Joint Learning}
In this learning framework, the gradient cannot be back-propagated from the representation module to the induction module due to the usage of $\mathop{\arg\max}$ operator. It is therefore desirable to connect these two modules in a way such that they can improve each other by their own estimations.
In one direction, forwarding the prediction of the sense induction module to the sense representation learning module is trivial, while in another direction, we treat the estimated collocation likelihood as the reward for the induction module.

First note that calculating the partition function in the denominator of (\ref{eq:4}) and (\ref{eq:5}) is intractable since it involves a computationally expensive summation over all sense identities.
In practice, we adopt the negative sampling strategy technique~\cite{mikolov2013distributed} and rewrite (\ref{eq:4}) and (\ref{eq:5}) as:

\begin{align}
    	\log p(s_j\mid s_i) &= \nonumber \log\sigma((U^{en}_{s_i})^TV^{en}_{s_j})+\\
        &\sum_{k=1}^{N} \mathbb{E}_{s_k\sim p_\text{neg}(s)}[\sigma(-(U^{en}_{s_i})^TV^{en}_{s_k})], \label{eq:6}\\
    \log p(s'_l\mid s_i) &= \nonumber \log\sigma((U^{en}_{s_i})^TV^{zh}_{s'_l})+\\
            &\sum_{k=1}^{N} \mathbb{E}_{s'_k\sim p_\text{neg}(s')}[\sigma(-(U^{en}_{s_i})^TV^{zh}_{s'_k})],\label{eq:7}
\end{align}
where $p_\text{neg}(s)$ and $p_\text{neg}(s')$ is the distribution over all English senses and all Chinese senses for negative samples respectively, and $N$ is the number of negative sample.
The rewritten objective for optimizing two sense representation learning modules is the same as maximizing
(\ref{eq:6}) and (\ref{eq:7}).
Moreover, we can utilize the probability of correctly classifying the skip-gram sense pair as the reward signal. 
The intuition is that a correctly decoded sense identity is more likely to predict its neighboring sense identity compared to incorrectly decoded ones.

This learning framework can now be viewed as a reinforcement learning agent solving one-step Markov Decision Process~\cite{sutton1998reinforcement,lee2017muse}. 
For bilingual modules, the state, action, and reward correspond to bilingual context $\bar{C}$, sense $z_{ik}$, and $\sigma((U^{en}_{s_i})^TV^{zh}_{s'_l})$ respectively.
As for the monolingual modules, the state, action, and reward correspond to monolingual context $c_t$, sense $z_{ik}$, and $\sigma((U^{en}_{s_i})^TV^{en}_{s_j}))$. 
Finally, we can optimize both bilingual and monolingual sense induction modules ($P$ and $Q$ from (\ref{eq:2})
by minimizing the cross entropy loss between decoded sense probability and reward. We also include an entropy regularization term as suggested in~\cite{vsuster2016bilingual} to let the sense induction module converge faster and make more confident predictions. Formally,

\begin{align}
	\min H(\sigma((U^{en}_{s_i})^TV^{zh}_{s'_l})&, p(z_{ik} \mid c_i, c_i'))\nonumber & \\ 
        &+ \lambda E(p(z_{ik} \mid c_i, c_i')) \label{eq:8}\\
    \min H(\sigma((U^{en}_{s_i})^TV^{en}_{s_j})&, p(z_{ik} \mid c_i)) &\nonumber \\
    	&+ \lambda E(p(z_{ik} \mid c_i))  \label{eq:9}
\end{align}
$E$ is the entropy of selection probability weighted by $\lambda$. Note that the major language is switched iteratively among two languages.
Algorithm~\ref{algo:learning} presents the full learning procedure.

\begin{algorithm}[t]
\caption{Bilingual Sense Embedding Learning Algorithm}\label{algo:learning}
\begin{algorithmic}[1]
\small
\Require $C^{en}$, $C^{zh}$, $W^{en}$, $W^{zh}$
\Ensure $P^{en}$, $P^{zh}$, $Q^{en}$, $Q^{zh}$, $U^{en}$, $U^{zh}$, $V^{en}$, $V^{zh}$
\Loop \text{ until converge}
    \State \Call{Main}{\textit{en}, \textit{zh}, 0.4}
    \Comment{0.4 is just an example weight}
    \State \Call{Main}{\textit{zh}, \textit{en}, 0.4}
\EndLoop

\Function{Main}{maj, bi, $\alpha$}
	\State $t, i, j, k, l \gets$ \Call{GetTrainData}{maj}
	\State $s_i, pred_i \gets$ \Call{InduceSense}{maj, bi, $t$, $i$, $\alpha$}
    \State $s_j, \_ \gets$ \Call{InduceSense}{maj, bi, $t$, $j$, $\alpha$}
    \State $s'_l, pred'_l \gets$ \Call{InduceSense}{bi, bi, $t$, $k$, 1.0}
    \State $s'_k, \_ \gets$ \Call{InduceSense}{bi, bi, $t$, $l$, 1.0}
    \State $r \gets$ \Call{TrainSRL}{maj, maj, $s_i$, $s_j$}
    \State $r' \gets$\Call{TrainSRL}{maj, bi, $s_i$, $s'_l$}
    \State $r'' \gets$\Call{TrainSRL}{bi, bi, $s'_l$, $s'_k$}
    \State \Call{TrainSI}{maj, bi, $r,pred_i$}
    \State \Call{TrainSI}{maj, bi, $r',pred_i$}
    \State \Call{TrainSI}{bi, bi, $r'',pred'_l$}
\EndFunction

\Function{InduceSense}{maj, bi, $t$, $i$, $\alpha$}
    \State \text {calculate $\alpha$-weighted $\bar{C}$ by (\ref{eq:1}) } 
    	\State \text{select $z_{ik}^*$ by (\ref{eq:2}) and (\ref{eq:3})}
    \State \Return {$z_{ik}^*$, $p(z_{ik}^* \mid \bar{C})$}
\EndFunction

\Function{TrainSRL}{maj, bi, $s_i$, $s_j$}

\If {maj==bi}
	\State \text {optimize $U^{maj}$, $V^{maj}$ by (\ref{eq:6}) given $s_i$, $s_j$}
\Else
	\State \text {optimize $U^{maj}$, $V^{bi}$ by (\ref{eq:7}) given $s_i$, $s_j$}
\EndIf
\State \Return collocation prob of $(s_i, s_j)$%
\EndFunction

\Function{TrainSI}{maj, bi, r, pred}
\If {maj==bi}
	\State \text {optimize $P^{maj}$, $Q^{maj}$ by (\ref{eq:9}) given r, pred}
\Else
	\State \text {optimize $P^{maj}$, $Q^{bi}$ by (\ref{eq:8}) given r, pred}
\EndIf

\EndFunction
\end{algorithmic}
\end{algorithm}

\begin{table*}
\centering
\begin{tabular}{llc}
\hline
\bf English Sentence & \bf Chinese Sentence & \bf Score\\ 
\hline
	Judges must give both sides an equal  
	&\begin{CJK}{UTF8}{bkai}我非常喜歡這個故事，它\textbf{$<$告訴$>$}我們一些\end{CJK} & 7.00\\
	opportunity to \textbf{$<$state$>$} their cases.
    &\begin{CJK}{UTF8}{bkai}重要的啟示。\end{CJK} (I like this story a lot, which\\
    & \textbf{$<$tells$>$} us some important inspiration.) \\
\hline
	It was of negligible \textbf{$<$importance$>$} prior  
	&\begin{CJK}{UTF8}{bkai}黃斑部病變的預防及早期治療是相當\textbf{$<$重要$>$}\end{CJK} & 6.94\\
	to 1990, with antiquated weapons and 
    &\begin{CJK}{UTF8}{bkai}的。\end{CJK} (The prevention and early treatment of  \\
    few members. & macular lesions is very \textbf{$<$important$>$}.)&  \\
\hline
	Due to the San Andreas Fault bisecting  
	&\begin{CJK}{UTF8}{bkai}水果攤老闆似乎很意外真有人買這\textbf{$<$冷$>$}貨\end{CJK} & 3.70\\
	 the hill, one side has \textbf{$<$cold$>$} water, the 
    &\begin{CJK}{UTF8}{bkai}，露出「你真內行」的眼神與我聊了幾句。\end{CJK} \\
    other has hot. & (The owner of the fruit stall seemed surprised \\
    &  that someone bought this \textbf{$<$unpopular$>$} product,\\ &talking me few words about ``you are such a pro''.)\\
\hline
\end{tabular}
\caption{\label{tb:bcws} Sentence pair examples and average annotated scores in BCWS.}
\end{table*}

\section{New Dataset---Bilingual Contextual Word Similarity (BCWS)}
We propose a new dataset to measure the bilingual contextual word similarity. English and Chinese are chosen as our language pair for three reasons:
\begin{compactenum}
    \item They are the top widely used languages in the world.
    \item English and Chinese belong to completely different language families, making it interesting to explore syntactic and semantic difference among them.
    \item Chinese is a language that requires segmentation, this dataset can also help researchers experiment on different segmentation levels and investigate how segmentation affects the sense similarity.
\end{compactenum}
This dataset also provides a \emph{direct} measure to determine whether the two language embeddings align well in the vector space.
Note that we focus on word-level, and this is different from~\cite{klementiev2012inducing}, which also measured the cross-lingual embedding similarity but rely on the ambiguous document-level classification.

Our dataset contains 2091 question pairs, where each pair consists of exactly one English and one Chinese sentence; note that they are~\textbf{not} parallel but with their own sentential contexts shown in Table~\ref{tb:bcws}.
Eleven raters\footnote{They are all Chinese native speaker whose scores are at least 29 in the TOEFL reading section or 157 in the GRE verbal section.} were recruited to annotate this dataset.
Each rater gives a score ranging from 1.0 (different) to 10.0 (same) for each question to indicate the semantic similarity of bilingual word pairs based on sentential clues.
The annotated dataset shows very high intra-rater consistency; we leave one rater out and calculate Spearman correlation between the rater and the average of the rest, and the average number is about 0.83, indicating the human-level performance (the average number in SCWS is 0.52).

We describe the construction of BCWS below.
\paragraph{Chinese Multi-Sense Word Extraction}
\label{ssec:ch_w}
We utilize the Chinese Wikipedia dump to extract the most frequent 10000 Chinese words that are \emph{nouns}, \emph{adjective}, and \emph{verb} based on Chinese Wordnet~\cite{huang2010chinese}.
In order to test the sense-level representations, we discard single-sense words to ensure that the selected words are polysemous.
Also, the words with more than 20 senses are deleted, since those senses are too fine-grained and even hard for human to disambiguate.
We denote the list of Chinese words $l_c$.

\paragraph{English Candidate Word Extraction}
We have to find an English counterpart for each Chinese word in $l_c$.
We utilize \emph{BabelNet}~\cite{navigli2010babelnet}, a free and open-sourced knowledge resource, to serve as our bilingual dictionary. 
To be more concrete, we first query the selected Chinese word using the free API call provided by Babelnet to retrieve all \textit{WordNet} senses\footnote{\emph{BabelNet} contains sense definitions from various resources such as Wordnet, Wikitionary, Wikidata, etc}.
For example, the Chinese word
``\begin{CJK}{UTF8}{bkai}制服\end{CJK}''
has two major meanings:
\begin{compactitem}
\vspace{2mm}
	\item \textit{a type of clothing worn by members of an organization}
	\item \textit{force to submit or subdue}.
\vspace{2mm}
\end{compactitem}
Hence, we can obtain two candidate English words ``\textit{uniform}'' and ``\textit{subjugate}''.
Each word in $l_c$ retrieves its associated English candidate words and obtain the dictionary $D$.

\paragraph{Enriching Semantic Relationship}
Note that $D$ is merely a simple translation mapping between Chinese and English words.
It is desirable that we have a more complicated and interesting relationship between bilingual word pairs.
Hence, we traverse $D$ and for each English word we find its \emph{hyponyms}, \emph{hypernyms}, \emph{holonyms} and \emph{attributes}, and add the additional words into $D$.
In our example, we may obtain \{\begin{CJK}{UTF8}{bkai}制服\end{CJK}:[uniform, subjugate, livery, clothing, repress, dominate, enslave, dragoon...]\}.
We sample 2 English words if the number of English candidate words is more than 5, 3 English words if more than 10, and 1 English word otherwise to form the final bilingual pair.
For example, a bilingual word pair (\begin{CJK}{UTF8}{bkai}制服\end{CJK}, enslave) can be formed accordingly.
After this step, we obtain 2091 bilingual word pairs $P$.

\paragraph{Adding Contextual Information}
Given the bilingual word pairs $P$, appropriate contexts should be found in order to form the full sentences for human judgment.
For each Chinese word, we randomly sample one example sentence in Chinese WordNet that matches the PoS tag we selected in section~\ref{ssec:ch_w}.
For each English word, we traverse the whole English Wikipedia dump to find the sentences that contain the target English word.
We then sample one sentence where the target word is tagged as the matched PoS tag\footnote{We use the NLTK PoS tagger to obtain the tags.}.

\section{Experiments}
\subsection{Experimental Setup}
Two sets of parallel data are used in the experiments, one for English-Chinese (EN-ZH) and another for English-German (EN-DE).
UM-corpus~\cite{tian2014corpus} is used for EN-ZH training, while Europarl corpus~\cite{koehn2005europarl} is used for EN-DE training.
UM-corpus contains 15,764,200 parallel sentences with 381,921,583 English words and 572,277,658 unsegmented Chinese words. 
Europarl contains 1,920,209 parallel sentences with 44,548,491 German words and 47,818,827 English words. 
We evaluate our proposed model on the benchmark monolingual dataset, SCWS, and on the bilingual dataset, our proposed BCWS, where the evaluation metrics are actually introduced in section~\ref{sec:metric}.

\subsection{Hyperparameter Settings}
In our experiments, we use a mini-batch size of 512, context window size for major language is set to $m=5$ and we sample $M=20$ words for bilingual context. For the exploration of sense induction module, we set $\epsilon=0.05$. The $\lambda$ of entropy regularization is set to 1.\footnote{We tried different values of $\lambda=0.001, 0.5$, and the model converges approximately 12, 5 times slower compared to $\lambda=1$.} For negative sampling in (\ref{eq:6}) and (\ref{eq:7}), we pick $N=25$. The fixed learning rate is set to 0.025. The embedding dimension is 300 and the sense number per word is set to 3 for both Chinese, German, and English ($|Z_i|=3$). This setting is for a fair comparison with prior works.

\begin{table*}
  \centering
  \begin{tabular}{lcccccccc}
    \hline
    \multirow{2}{*}{\bf Model} &  \multirow{2}{*}{\bf $\alpha$} & \multicolumn{2}{c}{\bf EN-ZH} & \bf EN-DE\\
    \cline{3-4}
    & & \bf Bilingual/BCWS & \bf Mono(EN)/SCWS & \bf Mono(EN)/SCWS\\
    \hline
    \multicolumn{2}{l}{\it 1) Monolingual Sense Embeddings}\\
    \citet{lee2017muse} & & & {\bf66.8} / 65.5 & 63.8 / 63.4\\ 
    \hline
    \multicolumn{2}{l}{\it 2) Cross-Lingual Word Embeddings}\\
    \citet{luong2015bilingual} & & 50.4 & 61.1 & 62.1 \\
    \citet{conneau2017word} & & 54.7 & 65.5 & 64.0 \\
    \hline
    \multicolumn{2}{l}{\it 3) Cross-Lingual Sense Embeddings}\\
    \citet{upadhyay2017beyond} & & - & 45.0$^\star$ & - \\
    Proposed & 0.1 & 58.3 / 58.3 & 65.8 / 65.8 & 63.1 / 63.3 \\
    & 0.3 & \bf 58.8 / 58.8 & 65.9 / 66.0 & 63.5 / 63.9 \\
    & 0.5 & 58.5 / 58.5 & 66.7 / \bf 67.0 & 63.7 / 64.3 \\
    & 0.7 & 58.3 / 58.4 & 66.3 / 66.6 & 63.7 / 64.1 \\
    & 0.9 & 58.3 / 58.3 & 66.1 / 66.2 & \bf 63.9 / 64.6 \\
    \hline
  \end{tabular}

  \caption{Contextual similarity results evaluated on the SCWS/BCWS dataset, where the reported numbers indicate Spearman's rank correlation $\rho\times 100$ on AvgSimC / MaxSimC.$^\star$ indicates that \citet{upadhyay2017beyond} trained the sense embeddings using a different parallel dataset.}
  \label{tb:all}
\end{table*}

\subsection{Baseline}
The baselines for comparison can be categorized into three:
\begin{compactitem}
\item {\it Monolingual sense embeddings}:
\citet{lee2017muse} is the current state-of-the-art model of monolingual sense embedding evaluated on SCWS.
We re-train the sense embeddings using the same data but only in English for  fair comparison.
\item {\it Cross-lingual word embeddings}:
\citet{luong2015bilingual} treated words from different languages the same and trained cross-lingual embeddings in the same space.
\citet{conneau2017word} utilized adversarial training to map pretrained word embeddings into another language space.
\item {\it Cross-lingual sense embeddings}:
\citet{upadhyay2017beyond} utilized more than two languages to learn multilingual embeddings.
We report the number shown in the paper for comparison.
\end{compactitem}

\subsection{Evaluation Metric}
\label{sec:metric}
\citet{reisinger2010multi} introduced two contextual similarity estimations, AvgSimC and MaxSimC.
AvgSimC is a \emph{soft} measurement that addresses the contextual information with a probability estimation:
\begin{align}
\text{AvgSimC} & (w_i, \bar{C_t}, w_j, \bar{C_{t'}}) = \nonumber\\
\sum_{k=1}^{|Z_i|} & \sum_{l=1}^{|Z_j|} \pi(z_{ik}|\bar{C_t}) \pi(z_{jl}|\bar{C_{t'}}) d(z_{ik}, z_{jl}),\nonumber
\end{align}

AvgSimC weights the similarity measurement of each sense pair $z_{ik}$ and $z_{jl}$ by their probability estimations.
On the other hand, MaxSimC is a \emph{hard} measurement that only considers the most probable senses:
\vspace{-3pt}
\begin{align*}
\text{MaxSimC}&(w_i, \bar{C_t}, w_j, \bar{C_{t'}}) = d(z_{ik},z_{jl}),\\
z_{ik} = &\arg\max_{z_{ik'}}\pi(z_{ik'}|\bar{C_t}),\\
z_{jl} = &\arg\max_{z_{jl'}}\pi(z_{jl'}|\bar{C_{t'}}).\nonumber
\end{align*}
$d(z_{ik}, z_{jl})$ refers to the cosine similarity between $U^{maj}_{z_{ik}}$ and $U^{bi}_{z_{jl}}$ in the bilingual case (BCWS) and  $U^{maj}_{z_{ik}}$ and $U^{maj}_{z_{jl}}$ in the monolingual case (SCWS).

\subsection{Bilingual Embedding Evaluation}

Cross-lingual sense embeddings are the main contribution of this paper.
Table~\ref{tb:all} shows that all results from the proposed model are significantly better than the baselines that learn cross-lingual word embeddings.
It indicates that the sense-level information is critical for precise vector representations.
In addition, all results for AvgSimC and MaxSimC are the same in the proposed model, showing that the learned selection distribution is reliable for sense decoding.

\subsection{Monolingual Embedding Evaluation}
Because our model considers multiple languages and learns the embeddings jointly, the multilingual objective makes learning more difficult due to more noises.
In order to ensure the quality of the monolingual sense embeddings, we also evaluate our learned English sense embeddings on the benchmark SCWS data.
Comparing the results between training on EN-ZH and training on EN-DE, all results using EN-ZH are better than ones using EN-DE.
The probable reason is that the language difference between English and Chinese is larger than English and German; parallel Chinese sentences therefore provide informative cues for learning better sense embeddings.
Furthermore, our proposed model achieves comparable or superior performance than the current state-of-the-art monolingual sense embeddings proposed by \citet{lee2017muse} when trained on our monolingual data.

\begin{table}
  \centering
  \begin{tabular}{llcccccccc}
    \hline
    \bf Model    & \small \bf EN2DE & \small \bf DE2EN \\
    \hline
    \multicolumn{2}{l}{\it 1) Sentence-Level Training}\\
    \citet{hermann2014multilingual} & 83.7 & 71.4 \\
    \citet{ap2014autoencoder} &  \bf 91.8 & 72.8\\ 
    \citet{wei2017variational} & 91.0 & \bf 80.4\\ 
    \hline
    \multicolumn{2}{l}{\it 2) Word-Level Training}\\
    \citet{klementiev2012inducing} & 77.7 & 71.1\\ 
    \citet{gouws2015bilbowa} & 86.5 & 75.0\\ 
    \citet{kovcisky2014learning} & 83.1 & 75.4\\ 
    \citet{shi2015learning} & \bf 91.3 & \bf 77.2 \\ 
    \citet{luong2015bilingual} & 86.4 & 75.6\\ 
    \citet{conneau2017word} & 78.7 & 67.1 \\ 
    Proposed & 81.8 & 76.0\\
    \hline
  \end{tabular}
  \caption{ Accuracy on cross-lingual document classification (\%).}
  \label{tb:cldc}
\end{table}

\begin{table*}[t]
\centering
\begin{tabular}{lll}
\hline
\bf Target & \bf kNN Senses (EN) & \bf kNN Senses (ZH) \\
\hline
apple\_0 & fruit, cake, sweet & \begin{CJK}{UTF8}{bkai}蘋果, 春天, 蛋糕, \underline{iphone}, 雞蛋, 巧克力, 葡萄\end{CJK}\\ 
& & (apple, spring, cake, \underline{iphone}, egg, chocolate, purples)\\
apple\_1 & iphone, \underline{cake}, google, stores & \begin{CJK}{UTF8}{bkai}蘋果, iphone, 微軟, 競爭對手, \underline{春天}, 谷歌\end{CJK}\\
& & (apple, iphone, microsoft, competitor, \underline{spring}, google) \\
\hline
uniform\_0 & dressed, worn, tape, wearing, cloth & \begin{CJK}{UTF8}{bkai}\underline{均勻},光滑,衣服,鞋子,穿著,服裝\end{CJK}\\
& & (\underline{even}, smooth, clothes, shoes, wearing, clothing)\\
uniform\_1 & particle, computed, varying, gradient & \begin{CJK}{UTF8}{bkai}態,粉末,縱向,等離子體,剪切,剛度\end{CJK}\\
& & (phase, powder, longitudinal, plasma, cut, stiffness) \\
\hline
\end{tabular}
\caption{ Words with similar senses obtained by kNN.}
\label{tb:knn}
\end{table*}

\subsection{Sensitivity of Bilingual Contexts}
To investigate how much the bilingual sense induction module relies on another language, the results with different $\alpha$ are shown in the table.

To justify the usefulness of utilizing bilingual signal, we compare our model with~\citet{lee2017muse}, which used monolingual signal in a similar modular framework. Our method outperforms theirs in terms of MaxSimC on both EN-ZH and EN-DE. 
However, the performance is roughly the same on AvgSimC.
The reason may be that the bilingual signal is indicative but noisy, which largely affects AvgSimC due to its weighted sum operation. 
MaxSimC only picks the most probable senses, which makes it robust to noises.


In addition, our performance improves as $\alpha$ increases for EN-DE, and the best is obtained when $\alpha$ is large. This is interesting if we compare $\alpha=0.9$ to \emph{MUSE}, we can see that AvgSimC is similar but ours outperforms \emph{MUSE} on MaxSimC, indicating this little bilingual signal does help disambiguate senses more confidently.
In contrast, the best performance is obtained on EN-ZH when two languages have equal contribution. Because English is very different from Chinese, it can benefit more from Chinese than from German.

\subsection{Extrinsic Evaluation}
We further evaluate our bilingual sense embeddings using a downstream task, cross-lingual document classification (CLDC), with a standard setup~\cite{klementiev2012inducing}. 
To be more concrete, a set of labeled documents in language \emph{A} is available to train a classifier, and we are interested in classifying documents in another language \emph{B} at test time, which tests semantic transfer of information across different languages.
We use the averaged sense embeddings as word embeddings for a fair comparison.

The result is shown in Table~\ref{tb:cldc}. We can see that our proposed model achieves comparable performance or even superior performance to most prior work on the DE2EN direction;
however, the same conclusion does not hold for the EN2DE direction.
The reason may be that we test the model that works best on BCWS and hence not able to tune hyperparameters on the development set of CLDC.
In addition, we use the average of sense vectors as input word embeddings, which may induce some noises into the resulting vectors.
In sum, the comparable performance of the downstream task shows the practical usage and the potential extension of the proposed model.

\subsection{Qualitative Analysis}
Some examples of our learned sense embeddings are shown in Table~\ref{tb:knn}.
It is obvious to see that the first sense of~\textit{Apple} is related to~\textit{fruit and things to eat}, while the second one means the ~\textit{tech company Apple Inc}. 
Most English and Chinese nearest neighbors match the meanings of the induced senses, but there are still some noises that are underlined. 
For example,~\textit{cake} should be the neighbor of the first sense rather than the second one.
The same observation applies to~\textit{iphone} and~\textit{spring}. 
In our second example for \textit{uniform}, the first sense is related to \textit{outfit and clothes}, while the second is related to \textit{engineering terms}.
However, \textit{even} appears in the \textit{outfit and clothes} sense, which is incorrect.
The reason may be that the size of the parallel corpus is not large enough for the model to accurately distinguish all senses via unsupervised learning.
Hence, utilizing external resources such as bilingual dictionaries or designing a new model that can use existing large monolingual corpora like Wikipedia are our future work.

\section{Conclusion}
This paper is the first purely sense-level cross-lingual representation learning model with efficient sense induction, where several monolingual and bilingual modules are jointly optimized.
The proposed model achieves superior performance on both bilingual and monolingual evluation datasets.
A newly collected dataset for evaluating bilingual contextual word similarity is presented, which provides potential research directions for future work.

\section*{Acknowledgement}
We would like to thank reviewers for their insightful comments on the paper. 
This work was financially supported from the Young Scholar Fellowship Program by Ministry of Science and Technology (MOST) in Taiwan, under Grant 107-2636-E-002-004.

\newpage
\bibliographystyle{acl_natbib}
\bibliography{emnlp2018}

\begin{thebibliography}{28}
\expandafter\ifx\csname natexlab\endcsname\relax\def\natexlab#1{#1}\fi

\bibitem[{AP et~al.(2014)AP, Lauly, Larochelle, Khapra, Ravindran, Raykar, and
  Saha}]{ap2014autoencoder}
Sarath~Chandar AP, Stanislas Lauly, Hugo Larochelle, Mitesh Khapra, Balaraman
  Ravindran, Vikas~C Raykar, and Amrita Saha. 2014.
\newblock An autoencoder approach to learning bilingual word representations.
\newblock In \emph{Advances in Neural Information Processing Systems}, pages
  1853--1861.

\bibitem[{Bansal et~al.(2012)Bansal, DeNero, and Lin}]{bansal2012unsupervised}
Mohit Bansal, John DeNero, and Dekang Lin. 2012.
\newblock Unsupervised translation sense clustering.
\newblock In \emph{Proceedings of the 2012 Conference of the North American
  Chapter of the Association for Computational Linguistics: Human Language
  Technologies}, pages 773--782. Association for Computational Linguistics.

\bibitem[{Bartunov et~al.(2016)Bartunov, Kondrashkin, Osokin, and
  Vetrov}]{bartunov2016breaking}
Sergey Bartunov, Dmitry Kondrashkin, Anton Osokin, and Dmitry Vetrov. 2016.
\newblock Breaking sticks and ambiguities with adaptive skip-gram.
\newblock In \emph{Artificial Intelligence and Statistics}, pages 130--138.

\bibitem[{Camacho-Collados et~al.(2017)Camacho-Collados, Pilehvar, Collier, and
  Navigli}]{camacho2017semeval}
Jose Camacho-Collados, Mohammad~Taher Pilehvar, Nigel Collier, and Roberto
  Navigli. 2017.
\newblock Semeval-2017 task 2: Multilingual and cross-lingual semantic word
  similarity.
\newblock In \emph{Proceedings of the 11th International Workshop on Semantic
  Evaluation (SemEval-2017)}, pages 15--26.

\bibitem[{Conneau et~al.(2017)Conneau, Lample, Ranzato, Denoyer, and
  J{\'e}gou}]{conneau2017word}
Alexis Conneau, Guillaume Lample, Marc'Aurelio Ranzato, Ludovic Denoyer, and
  Herv{\'e} J{\'e}gou. 2017.
\newblock Word translation without parallel data.
\newblock \emph{arXiv preprint arXiv:1710.04087}.

\bibitem[{Gouws et~al.(2015)Gouws, Bengio, and Corrado}]{gouws2015bilbowa}
Stephan Gouws, Yoshua Bengio, and Greg Corrado. 2015.
\newblock Bilbowa: Fast bilingual distributed representations without word
  alignments.
\newblock In \emph{International Conference on Machine Learning}, pages
  748--756.

\bibitem[{Guo et~al.(2014)Guo, Che, Wang, and Liu}]{guo2014learning}
Jiang Guo, Wanxiang Che, Haifeng Wang, and Ting Liu. 2014.
\newblock Learning sense-specific word embeddings by exploiting bilingual
  resources.
\newblock In \emph{Proceedings of COLING 2014, the 25th International
  Conference on Computational Linguistics: Technical Papers}, pages 497--507.

\bibitem[{Hermann and Blunsom(2014)}]{hermann2014multilingual}
Karl~Moritz Hermann and Phil Blunsom. 2014.
\newblock Multilingual models for compositional distributed semantics.
\newblock \emph{arXiv preprint arXiv:1404.4641}.

\bibitem[{Huang et~al.(2010)Huang, Hsieh, Hong, Chen, Su, Chen, and
  Huang}]{huang2010chinese}
Chu-Ren Huang, Shu-Kai Hsieh, Jia-Fei Hong, Yun-Zhu Chen, I-Li Su, Yong-Xiang
  Chen, and Sheng-Wei Huang. 2010.
\newblock Chinese wordnet: Design, implementation, and application of an
  infrastructure for cross-lingual knowledge processing.
\newblock \emph{Journal of Chinese Information Processing}, 24(2):14--23.

\bibitem[{Huang et~al.(2012)Huang, Socher, Manning, and Ng}]{HuangEtAl2012}
Eric~H. Huang, Richard Socher, Christopher~D. Manning, and Andrew~Y. Ng. 2012.
\newblock {Improving Word Representations via Global Context and Multiple Word
  Prototypes}.
\newblock In \emph{Annual Meeting of the Association for Computational
  Linguistics (ACL)}.

\bibitem[{Klementiev et~al.(2012)Klementiev, Titov, and
  Bhattarai}]{klementiev2012inducing}
Alexandre Klementiev, Ivan Titov, and Binod Bhattarai. 2012.
\newblock Inducing crosslingual distributed representations of words.
\newblock \emph{Proceedings of COLING 2012}, pages 1459--1474.

\bibitem[{Ko{\v{c}}isk{\`y} et~al.(2014)Ko{\v{c}}isk{\`y}, Hermann, and
  Blunsom}]{kovcisky2014learning}
Tom{\'a}{\v{s}} Ko{\v{c}}isk{\`y}, Karl~Moritz Hermann, and Phil Blunsom. 2014.
\newblock Learning bilingual word representations by marginalizing alignments.
\newblock \emph{arXiv preprint arXiv:1405.0947}.

\bibitem[{Koehn(2005)}]{koehn2005europarl}
Philipp Koehn. 2005.
\newblock Europarl: A parallel corpus for statistical machine translation.
\newblock In \emph{MT summit}, volume~5, pages 79--86.

\bibitem[{Lee and Chen(2017)}]{lee2017muse}
Guang-He Lee and Yun-Nung Chen. 2017.
\newblock {MUSE}: Modularizing unsupervised sense embeddings.
\newblock In \emph{Proceedings of the 2017 Conference on Empirical Methods in
  Natural Language Processing}, pages 327--337.

\bibitem[{Li and Jurafsky(2015)}]{li2015multi}
Jiwei Li and Dan Jurafsky. 2015.
\newblock Do multi-sense embeddings improve natural language understanding?
\newblock \emph{Proceedings of the 2015 Conference on Empirical Methods in
  Natural Language Processing}, pages 1722--1732.

\bibitem[{Luong et~al.(2015)Luong, Pham, and Manning}]{luong2015bilingual}
Thang Luong, Hieu Pham, and Christopher~D Manning. 2015.
\newblock Bilingual word representations with monolingual quality in mind.
\newblock In \emph{Proceedings of the 1st Workshop on Vector Space Modeling for
  Natural Language Processing}, pages 151--159.

\bibitem[{Mikolov et~al.(2013)Mikolov, Sutskever, Chen, Corrado, and
  Dean}]{mikolov2013distributed}
Tomas Mikolov, Ilya Sutskever, Kai Chen, Greg~S Corrado, and Jeff Dean. 2013.
\newblock Distributed representations of words and phrases and their
  compositionality.
\newblock In \emph{Proceedings of Advances in neural information processing
  systems}, pages 3111--3119.

\bibitem[{Mnih et~al.(2013)Mnih, Kavukcuoglu, Silver, Graves, Antonoglou,
  Wierstra, and Riedmiller}]{mnih2013playing}
Volodymyr Mnih, Koray Kavukcuoglu, David Silver, Alex Graves, Ioannis
  Antonoglou, Daan Wierstra, and Martin Riedmiller. 2013.
\newblock Playing atari with deep reinforcement learning.
\newblock \emph{NIPS Deep Learning Workshop}.

\bibitem[{Navigli and Ponzetto(2010)}]{navigli2010babelnet}
Roberto Navigli and Simone~Paolo Ponzetto. 2010.
\newblock Babelnet: Building a very large multilingual semantic network.
\newblock In \emph{Proceedings of the 48th annual meeting of the association
  for computational linguistics}, pages 216--225. Association for Computational
  Linguistics.

\bibitem[{Neelakantan et~al.(2014)Neelakantan, Shankar, Passos, and
  McCallum}]{neelakantan2014efficient}
Arvind Neelakantan, Jeevan Shankar, Alexandre Passos, and Andrew McCallum.
  2014.
\newblock Efficient non-parametric estimation of multiple embeddings per word
  in vector space.
\newblock In \emph{Proceedings of the 2014 Conference on Empirical Methods in
  Natural Language Processing}.

\bibitem[{Reisinger and Mooney(2010)}]{reisinger2010multi}
Joseph Reisinger and Raymond~J Mooney. 2010.
\newblock Multi-prototype vector-space models of word meaning.
\newblock In \emph{Human Language Technologies: The 2010 Annual Conference of
  the North American Chapter of the Association for Computational Linguistics},
  pages 109--117. Association for Computational Linguistics.

\bibitem[{Shi et~al.(2015)Shi, Liu, Liu, and Sun}]{shi2015learning}
Tianze Shi, Zhiyuan Liu, Yang Liu, and Maosong Sun. 2015.
\newblock Learning cross-lingual word embeddings via matrix co-factorization.
\newblock In \emph{Proceedings of the 53rd Annual Meeting of the Association
  for Computational Linguistics and the 7th International Joint Conference on
  Natural Language Processing (Volume 2: Short Papers)}, volume~2, pages
  567--572.

\bibitem[{{\v{S}}uster et~al.(2016){\v{S}}uster, Titov, and van
  Noord}]{vsuster2016bilingual}
Simon {\v{S}}uster, Ivan Titov, and Gertjan van Noord. 2016.
\newblock Bilingual learning of multi-sense embeddings with discrete
  autoencoders.
\newblock In \emph{Proceedings of NAACL-HLT}, pages 1346--1356.

\bibitem[{Sutton and Barto(1998)}]{sutton1998reinforcement}
Richard~S Sutton and Andrew~G Barto. 1998.
\newblock \emph{Reinforcement learning: An introduction}, volume~1.
\newblock MIT press Cambridge.

\bibitem[{Tian et~al.()Tian, Wong, Chao, Quaresma, and
  Oliveira}]{tian2014corpus}
Liang Tian, Derek~F Wong, Lidia~S Chao, Paulo Quaresma, and Francisco Oliveira.
\newblock Um-corpus: A large english-chinese parallel corpus for statistical
  machine translation.

\bibitem[{Upadhyay et~al.(2017)Upadhyay, Chang, Taddy, Kalai, and
  Zou}]{upadhyay2017beyond}
Shyam Upadhyay, Kai-Wei Chang, Matt Taddy, Adam Kalai, and James Zou. 2017.
\newblock Beyond bilingual: Multi-sense word embeddings using multilingual
  context.
\newblock In \emph{Proceedings of the 2nd Workshop on Representation Learning
  for NLP}, pages 101--110.

\bibitem[{Wei and Deng(2017)}]{wei2017variational}
Liangchen Wei and Zhi-Hong Deng. 2017.
\newblock A variational autoencoding approach for inducing cross-lingual word
  embeddings.
\newblock In \emph{Proceedings of the 26th International Joint Conference on
  Artificial Intelligence}, pages 4165--4171. AAAI Press.

\bibitem[{Yarowsky(1993)}]{yarowsky1993one}
David Yarowsky. 1993.
\newblock One sense per collocation.
\newblock In \emph{Proceedings of the workshop on Human Language Technology},
  pages 266--271. Association for Computational Linguistics.

\end{thebibliography}


\end{document}